\documentclass{article}

\usepackage[preprint]{corl_2026}
\usepackage{graphicx}
\usepackage{booktabs}
\usepackage{xcolor}
\usepackage{siunitx}
\usepackage{multirow}

\title{A Head Camera Is (Almost) All You Need:\\An Electronics-Free UMI Glove Labelled and Observed by One Moving Camera}

\author{
  Yosub Shin \\
  University of Hawaii at Manoa \\
  \texttt{yosubs@hawaii.edu} \\
  \And
  Yuan Liu \\
  Independent \\
  \texttt{hiveyuan@gmail.com} \\
  \And
  Igor Molybog \\
  University of Hawaii at Manoa \\
  \texttt{molybog@hawaii.edu} \\
}

\begin{document}
\maketitle

\begin{abstract}
Handheld demonstration devices decouple data collection from the robot, but conventional designs burden the hand with a wrist camera, cables and a pose tracker (SLAM, an IMU, a VR controller, or a fixed external camera that the collector's body occludes), and burden the pipeline with as many data streams to timestamp, synchronise and cross-calibrate. We show that a single egocentric camera (\$122) worn on the collector's head replaces both the camera and the tracker, and the streams with one. It reads printed fiducials on an electronics-free glove (\$22) against paper tiles taped around the workspace, giving the 6-DoF gripper pose and the jaw width from the same frames the policy will see; at deployment the same camera rides a low-cost (\$120) arm beside the robot that mimics the collector's head from the robot's joint state. Against the robot's encoders the labels are accurate to \SI{6.5}{mm} and \SI{1.6}{\degree} median with \SI{93}{}--\SI{100}{\percent} frame coverage, and a policy trained on them instead of encoders pays \SI{3}{\percent} in validation loss. Adding 194 glove demonstrations to 48 teleoperated episodes lifts rollout success from 10/25 to 17/25, level with 172 teleoperated episodes alone (18/25), and the full co-trained head-camera policy succeeds in 20/25 against 15/25 for the wrist-camera recipe, whose policies drop the object when the goal leaves their view, and 12/25 for a fixed camera.
\end{abstract}

\begin{figure}[!h]
  \centering
  \includegraphics[width=\linewidth]{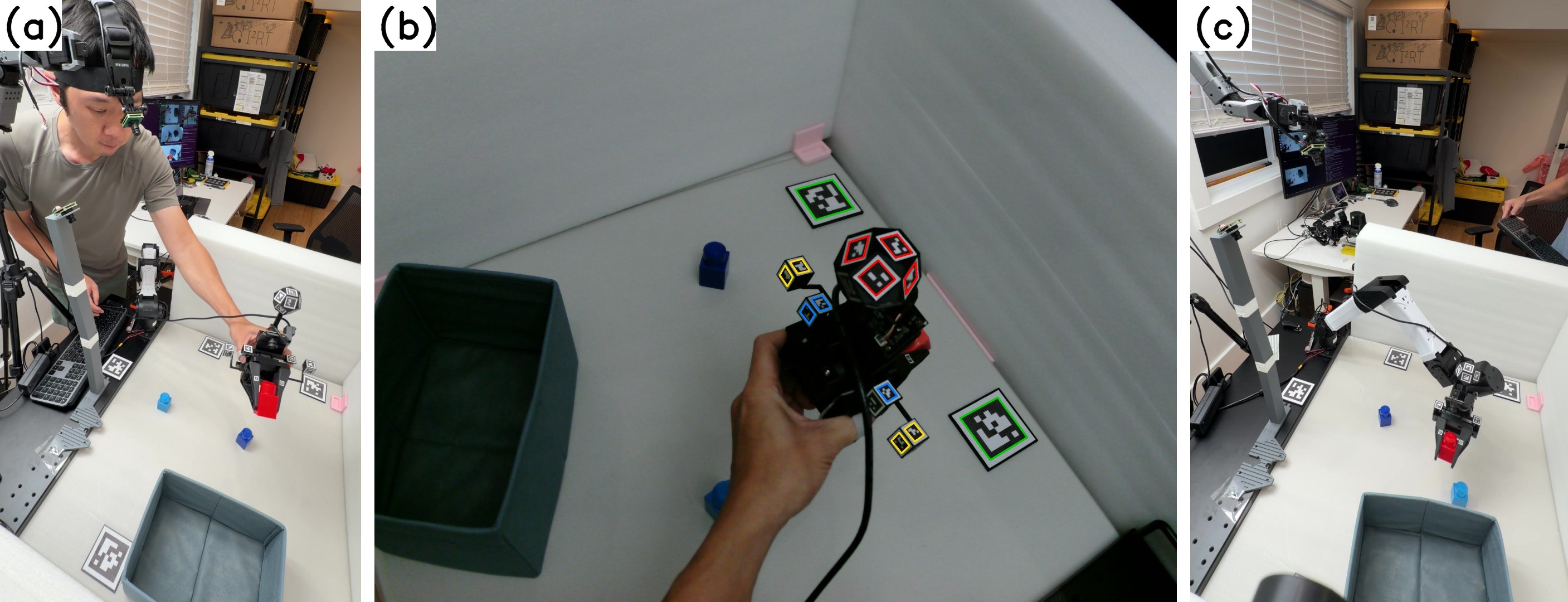}
  \caption{One camera, two mounts. (a) The collector wears the camera and holds the electronics-free glove over the tiled workspace. (b) The camera's own frame: tiles (green), ball (red) and jaw tags (blue, yellow) give the pose and the jaw width. (c) The same camera on a six-joint positioner beside the robot during a rollout. The wrist fisheye on the glove and the fixed camera on the post serve the ablations only.}
  \label{fig:teaser}
\end{figure}

\section{The idea}
\label{sec:idea}
Teleoperation occupies a robot and an operator for every demonstration. Handheld devices decouple collection from the robot: with ours a collector records 2.0--3.1 demonstrations per minute against 1.6 by teleoperation on the same task (\S\ref{sec:system}), in parallel and without robot time, with a \$22 printed glove, a \$122 wearable camera and four paper tiles. The glove is purely mechanical, a replica of the robot's gripper carrying fiducials and nothing else, and the one head-worn camera is at once the policy's observation and the source of the gripper's pose and jaw width, so there is no second sensor stream to synchronise or calibrate against.

Every UMI-style device answers two questions: \emph{what does the policy see}, and \emph{how is the gripper tracked}? The conventional answer puts both on the wrist. SLAM on the wrist camera \citep{chi2024umi,campos2021orbslam3} needs a mapping pass and textured scenes and loses track under occlusion or fast motion; a visual-inertial sensor adds a second device \citep{fastumi2024}; a VR controller adds weight (\SI{119}{g} of YUBI's \SI{319}{g} \citep{yubi2026,tinyumi2026}) and external tracking infrastructure \citep{vrtracking2024,quest2rehab2022}; a fixed overhead camera \citep{yamumi2026} avoids drift but ties collection to a mount and loses the hand behind the body in \SI{8}{}--\SI{22}{\percent} of frames (\S\ref{sec:labels}). Each tracker is a second stream beside the camera, with its own clock and calibration to reconcile, and the wrist camera itself adds cable and weight while its view is tied to the hand: even through UMI's \SI{155}{\degree} fisheye the frame holds what the jaws face, so a drop-off bin leaves it once the hand lifts or turns \citep{hsu2022hands}.

We answer both with one head-worn camera, addressing two challenges this workshop names \citep{umiarena2026}: tracking without SLAM and the human-to-robot viewpoint gap. Gaze precedes the hand \citep{land2001eye}, so the head keeps the gripper at arm's length, unoccluded by the torso, with the workspace around it. From that vantage an 11-face fiducial ball gives the pose, tags on the jaws the opening, and four AprilTag tiles \citep{olson2011apriltag} the world frame. At deployment the camera rides a small printed servo arm beside the robot that imitates head motion from the robot's joint state, identically in teleoperation and rollouts, so glove, teleoperated and rollout data share one viewpoint. EgoMimic \citep{kareer2025egomimic} co-trains on egocentric human video with the hand pose estimated from the video; our collector holds a replica of the robot's gripper, so contact and jaw width are the robot's. Active-vision teleoperation \citep{chuang2025avaloha,cheng2024opentelevision} steers a robot camera from a VR headset; our positioner needs neither operator nor headset.

\section{System}
\label{sec:system}

\paragraph{Collection.}
The glove is an unpowered printed replica of the YAM gripper \citep{i2rtyam} from \citet{yamumi2026}, stock fingers on a rack-and-pinion, with an 11-face \SI{15}{mm} ArUco dodecahedron \citep{garrido2014aruco,wu2017dodecapen} on the back of the hand. Fingertip tags facing a wrist camera \citep{chi2024umi} are edge-on from the head, so each jaw carries a printed wedge, the \emph{tail}, with \SI{12}{mm} AprilTags on five faces beside a fixed tag on the gripper back: one face is always visible, and the tail-to-back distance along the jaw axis gives the width. Parts cost \$22. The collector wears an Arducam B0587 4K camera (\$122) on a \$10 GoPro head strap (\SI{103}{g} worn), recording at \SI{15}{Hz} with a short exposure, aimed once per session so the frame centre coincides with the collector's own (App.~\ref{app:details}). Four \SI{60}{mm} AprilTag tiles define the workspace frame, placed where the gaze lands during grasping so that two stay in view, with no other tag in the scene. Continuous recording with keypress episode boundaries gives 2.0--3.1 pick-and-place demonstrations per minute (median \SI{6}{}--\SI{8}{s}) against 1.6 by teleoperation on the same arm (median \SI{14}{s}).

\paragraph{Labels from the head camera.}
\label{sec:labels}
Each frame yields (i) the camera pose from the tiles, (ii) the gripper pose in the camera frame from PnP over the visible ball faces, and (iii) the jaw width from the tails by a per-unit linear map (\SI{0.9}{}--\SI{3.8}{mm} holdout RMS on calipered spacers). Composing (i) and (ii) gives the tool pose in the tile frame; a one-time link to the robot base, fitted with the ball mounted on the YAM (App.~\ref{app:details}), turns it into joint-space actions comparable to teleoperation. Two mechanisms keep the camera localised under head motion: the tile layout is refit from the multi-tile frames of the footage itself (four-tile reprojection \SI{3.4}{px} to \SI{0.5}{px} RMS), and frames with a single tile are chained to the previous accepted pose under velocity bounds rather than discarded. Dropouts over \SI{0.3}{s} split an episode and segments under \SI{1}{s} are dropped: \SI{0.7}{}--\SI{7.4}{\percent} of frames are cut per session, 200 segments from 198 demonstrations. A fixed external camera on the same demonstrations loses \SI{8}{}--\SI{22}{\percent} to body occlusion and fragments them into 259 segments.

\paragraph{The robot's head.}
A six-joint SO-101 follower arm \citep{so101} (six Feetech STS3215 servos, \$122 in parts) stands beside the YAM and carries the same camera on a GoPro-pattern mount in place of its gripper (Fig.~\ref{fig:teaser}c). Its kinematics are fitted once from 60 tile-verified poses (\SI{7}{mm} / \SI{0.7}{\degree} residual). At each session start it parks at the collector's median head pose and registers against the tiles; thereafter it tracks the gripper from the YAM's joint encoders alone, transparent to the policy: a fixed-gain controller holds the gripper's bearing and a ridge-regression residual fitted on about 100 glove episodes reproduces head motion relative to the hand (\SI{46}{mm} / \SI{5.4}{\degree} median head-pose error on held-out episodes against \SI{87}{mm} / \SI{7.0}{\degree} for a static camera; App.~\ref{app:details}). The camera follows the gripper with \SI{0.4}{s} lag and \SI{16}{mm} median tile-referenced error, and frames it where the collector did (normalised gripper centroid 0.37/0.62 against 0.38/0.60).

\paragraph{Datasets and training.}
Teleoperation under the positioner gave 178 episodes (38.6k frames; 6 held out for validation); four glove sessions gave 198 demonstrations, 200 segments (50.0k frames; 6 held out). All policies are ACT \citep{zhao2023act} in LeRobot \citep{lerobot2024}, 50k steps, joint-space actions, best checkpoint by validation loss on the held-out teleoperated episodes. Following \citet{yamumi2026}, glove timelines are resampled to teleoperation pace and widths mapped into the gripper command space, with the closed state anchored to the teleoperators' minimum commanded squeeze (0.22 of stroke rather than the 0.25 median): the head-view policies released no block during the carry in 120 rollouts.

\section{Experiments}
\label{sec:results}

\paragraph{Design.}
Eight policies (Table~\ref{tab:arms}) span three views: the head camera, a wrist fisheye and a fixed external camera. Teleoperated actions come from the encoders except in arm 6, and glove labels from the head camera unless stated. The teleoperator watched the head-camera preview and all three cameras recorded every episode, so the arms share demonstration trajectories and differ only in observation and data composition. Arms 1--4 test co-training at two teleoperation budgets (172 and 48 episodes); arm 6 tests head-camera labels in place of encoders; arms 5, 7 and 8 test the view: arm 5 is the wrist-camera UMI recipe \citep{chi2024umi} with a fisheye added to the glove for this ablation only, arm 7 a fixed camera (\SI{10}{\percent} fewer glove frames to body occlusion), and arm 8 head labels on the wrist view, separating the view from the labels.

\begin{table}[t]
  \centering
  \footnotesize
  \caption{The eight policies. Teleop labels are encoders except arm 6. Val = ACT loss on 6 held-out teleoperated episodes with encoder targets. Rollouts: 25 placements, one trial each, \SI{60}{s} budget; $n$ = no grasp, $d$ = dropped before the bin, $b$ = dropped at the bin, $t$ = timeout.}
  \label{tab:arms}
  \setlength{\tabcolsep}{4pt}
  \begin{tabular}{clllcccl}
    \toprule
    \# & Arm & Glove labels & Observation & Teleop + glove & Val & Success & Fails \\
    \midrule
    1 & teleop only & --- & head cam & 172 & 0.318 & 18/25 & n5 d1 t1 \\
    2 & \textbf{device} & head cam & head cam & 172 + 194 & \textbf{0.263} & \textbf{20/25} & n5 \\
    3 & teleop only, low budget & --- & head cam & 48 & 0.464 & 10/25 & n15 \\
    4 & device, low budget & head cam & head cam & 48 + 194 & 0.295 & 17/25 & n8 \\
    5 & wrist-camera UMI & fixed cam & wrist fisheye & 168 + 253 & 0.257 & 15/25 & n6 d3 b1 \\
    6 & head-cam teleop labels & --- & head cam & 172 (head labels) & 0.328 & 15/25 & n10 \\
    7 & fixed camera as system & fixed cam & fixed cam & 172 + 255 & 0.256 & 12/25 & n12 b1 \\
    8 & head labels, wrist view & head cam & wrist fisheye & 168 + 192 & 0.252 & 14/25 & n3 d7 b1 \\
    \bottomrule
  \end{tabular}
\end{table}

\paragraph{Label accuracy.}
With the ball on the YAM and the encoders as reference, the chain places the tool point within \SI{6.5}{mm} median (\SI{14.1}{mm} p90) and \SI{1.55}{\degree} (\SI{2.70}{\degree}) over 5\,065 frames. The jaw width, checked on 112 teleoperated episodes (24.6k frames), is within \SI{3.5}{\percent} of stroke at the median (\SI{7.9}{\percent} p90), \SI{2.3}{\percent} in the closed state that decides a grasp, with no lag and a label on every frame in 9 of 10 episodes; the error is flat over workspace, distance, elevation and viewing angle (\SI{2.8}{}--\SI{4.8}{\percent} median per bin; App.~\ref{app:width}). A policy trained on these labels instead of encoders (arm 6 against arm 1), scored on the same encoder-labelled held-out episodes, loses \SI{3}{\percent} in validation loss (0.328 against 0.318) and is within noise on the robot (15/25 against 18/25; 5 placements won to 8 lost, exact McNemar $p = 0.58$).

\paragraph{Co-training and the teleoperation budget.}
Co-training across sources has precedent \citep{fu2024mobilealoha,kareer2025egomimic}. Adding the 194 glove segments to 172 teleoperated episodes lowers the robot-validation loss from 0.318 to 0.263 (arm 2 against 1); added to 48 episodes, from 0.464 to 0.295 (arm 4 against 3). By log-linear interpolation of the teleop-only curve (Fig.~\ref{fig:budget}) the glove set is worth 65--100 teleoperated episodes, 2--3 glove demonstrations per teleoperated one, and the 48-episode mix lands at the level of 96--172 teleoperated episodes; the gain shrinks with budget (0.075 at 24 episodes, 0.011 at 96), so glove data stretches a small teleoperated set rather than replacing the last few dozen. On the robot the 48-episode mix succeeds in 17/25 against 10/25 for 48 teleoperated episodes alone (9 placements won to 2 lost, $p = 0.065$) and 18/25 for all 172 (4 to 5). Every failure of both 48-episode arms is a missed grasp; the difference is recovery: in scene 5 the mix missed, backed off and re-grasped where the teleop-only policy never closed in.

\paragraph{Which camera.}
Offline the four co-trained views (arms 2, 5, 7, 8) sit within 0.011 of each other (0.252--0.263), and offline loss is a weak proxy for rollouts \citep{mandlekar2021robomimic}. The rollouts separate them: head camera 20/25, wrist fisheye 15/25, fixed camera 12/25, head labels on the wrist view 14/25. On shared placements the head camera wins 6/1, 11/3 and 8/2 against them ($p = 0.13$, $0.06$, $0.11$), so no pair clears $0.05$ at 25 placements, but no arm beats arm 2 on more than three. The failure modes split by view: head-camera policies fail by never grasping, wrist-camera policies by dropping the block mid-carry (12 of the 21 failures of arms 5 and 8, against 1 of 45 for the head-camera arms, a slip at lift-off). In the operator's notes every wrist-view drop came while the fisheye looked away from the bin and none with the bin in view (Fig.~\ref{fig:rollouts}); the head camera keeps gripper and bin in one frame through the carry. Same view, other labels (arm 8 against 5): 14/25 against 15/25, indistinguishable. Same labels, other view (8 against 2): 20/25 falls to 14/25 and the no-grasps become drops. The fixed camera (arm 7) is the weakest and fails like the head view, by not grasping (12 of 13); it was mounted low because the positioner took the high vantage point, so its \SI{10}{\percent} occlusion is an upper bound. Median time to success is \SI{16}{}--\SI{19}{s} for every arm: the views differ in whether the task completes, not in how fast.

\section{Limitations and what is next}
\label{sec:discussion}
The tile layout is refined by median single-tile poses rather than a full bundle adjustment \citep{triggs2000bundle}, which should tighten it further. A label needs a tile in view as well as the glove, so coverage depends on where the tiles are placed; we followed the placement rules of \S\ref{sec:system} but tried only one layout. The positioner lags \SI{0.4}{s} and hits joint limits on wide sweeps; its viewpoint model is a linear fit on one session with no tuning against rollouts, and the jaws were often occluded by the robot's wrist-camera bracket, which the head-camera system does not need and which stayed on only for the wrist-view ablations. Removing it is the cheapest next gain. The head camera sees the collector's forearm and fingers, so glove-only policies do not transfer (validation 0.52 against 0.26 co-trained), as with the fixed-camera device. Controls not run: a higher fixed camera; teleoperation under a wrist-camera preview, since the operator watched the head view; a second seed, so the paired tests sit at $p \geq 0.06$ with a consistent ranking; and a second collector, as one person with one glove recorded every session. A workspace away from the robot needs no robot-side calibration: the tile-to-base link, fitted here with the robot holding the ball, can be left free per session as UMI-style pipelines do with relative trajectories; the collector only needs tiles taped down with one, preferably two, in view. Nothing is single-handed: a second glove with its own tag ids is read by the same camera, and one positioner serves both arms. Hardware \citep{yamumi2026} and tools \citep{forwardcamumi2026} are MIT-licensed; the robot-side calibration and dataset tools are not yet released. Next are the bimanual version, multi-tile workspaces, in-the-wild sessions and a second task.

\bibliography{refs}

\appendix
\setcounter{figure}{1}
\renewcommand{\thefigure}{A\arabic{figure}}
\setcounter{figure}{0}

\section{Details}
\label{app:details}

\paragraph{Camera mounting and aiming.}
The Arducam B0587 sits on a \$10 GoPro head strap through a printed GoPro-pattern adapter (\SI{103}{g} worn, plus \SI{37}{g} of USB cable). With the live preview open, the collector looks at a point \SI{0.5}{}--\SI{1}{m} away and tilts the mount until the centre of the frame coincides with the centre of their own view, so the camera looks where the head looks. The positioner carries the same pattern: the SO-101 follower's gripper bracket is removed and its gripper servo reused as a wrist-yaw joint, so the one camera moves between head and robot without re-mounting.

\paragraph{Calibration.}
With the ball mounted on the YAM's gripper, the flange-to-ball link is solved as a robot-world hand-eye problem (OpenCV's \texttt{calibrateRobotWorldHandEye}, \citet{shah2013robotworld}), and the tile-world-to-base link is a trimmed Kabsch fit of the head-camera tool point (tiles $\times$ ball) to forward kinematics over one 5\,065-frame session (\SI{91}{\percent} inliers). The tile refit reduces four-tile reprojection error from \SI{3.4}{px} to \SI{0.5}{px} RMS and the disagreement between single-tile solves of the same frame from \SI{10}{}--\SI{13}{mm} to \SI{4}{}--\SI{9}{mm}. Single-tile frames are chained under velocity bounds of \SI{0.6}{m/s} and \SI{120}{\degree/s}.

\paragraph{Positioner control.}
Forward kinematics gives the gripper position; a fixed-gain controller holds its initial visual bearing, and the residual model adds the head's displacement relative to the hand. The model is a ridge regression on nine pose features plus the grasp bearing, a bearing history and the distance, fitted on about 100 glove episodes of one session; on 15 held-out episodes it predicts the collector's head pose to \SI{46}{mm} / \SI{5.4}{\degree} median (\SI{81}{mm} / \SI{9.6}{\degree} p90), against \SI{87}{mm} / \SI{7.0}{\degree} for a camera held at the median head pose. Commands are low-pass filtered, rate-limited to \SI{50}{\degree/s} and bounded to a safe volume.

\section{Jaw-width accuracy by condition}
\label{app:width}
On 112 teleoperated episodes (24.6k frames) the head-camera width is compared with the encoder value. Binned by workspace quartile, by distance to the camera (\SI{0.4}{}--\SI{0.8}{m}), by the camera's elevation (\SI{40}{}--\SI{87}{\degree}) and by the angle between the line of sight and the jaw axis, every bin reads \SI{2.8}{}--\SI{4.8}{\percent} of stroke median and \SI{6.8}{}--\SI{8.5}{\percent} p90, worst when the camera looks across the jaws rather than along them. Shifting the label by one frame in either direction raises the error, so there is no lag. The residual is a systematic under-read of \SI{4}{}--\SI{6}{\percent} at full opening, a scale error of the per-session fit (alternate-frame holdout) rather than noise. The fixed camera of \citet{yamumi2026} read the same gripper at \SI{4.4}{\percent} median with \SI{94}{\percent} coverage.

\section{Rollout protocol and figures}
\label{app:rollouts}
Each of the 25 placements was set once and re-placed to a saved reference image for every arm (ghost alignment), with the arm order shuffled per scene and a \SI{60}{s} budget; the operator scored success, no grasp, dropped before the bin, dropped at the bin or timeout. In scene 5, arm 8 released the block mid-carry with the block outside its view, as if it no longer registered the object in hand; arm 5 shows the same pattern in its three drops. Head-camera arms that fail do so before contact, stopping short of the block or closing beside it, and then hold until the budget expires.

\begin{figure}[h]
  \centering
  \includegraphics[width=0.6\linewidth]{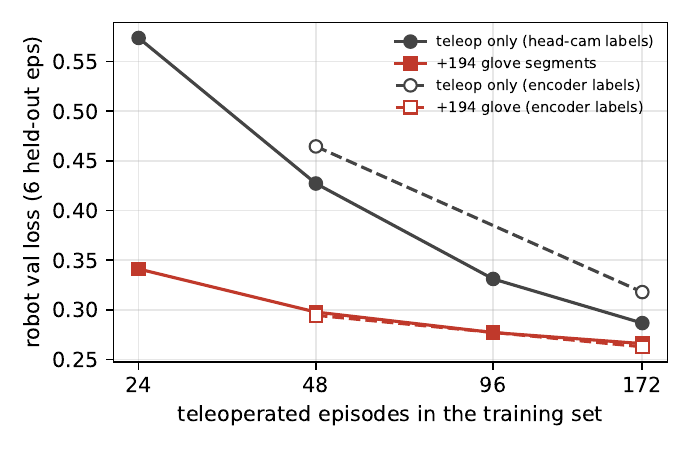}
  \caption{Robot validation loss against the number of teleoperated episodes, with and without the 194 glove segments. Solid: head-camera-labelled teleoperation (round-2 chain, four budgets); dashed: encoder-labelled teleoperation (final design, two budgets). The glove set moves every budget to the level of 2--4$\times$ as many teleoperated episodes.}
  \label{fig:budget}
\end{figure}

\begin{figure}[h]
  \centering
  \includegraphics[width=0.58\linewidth]{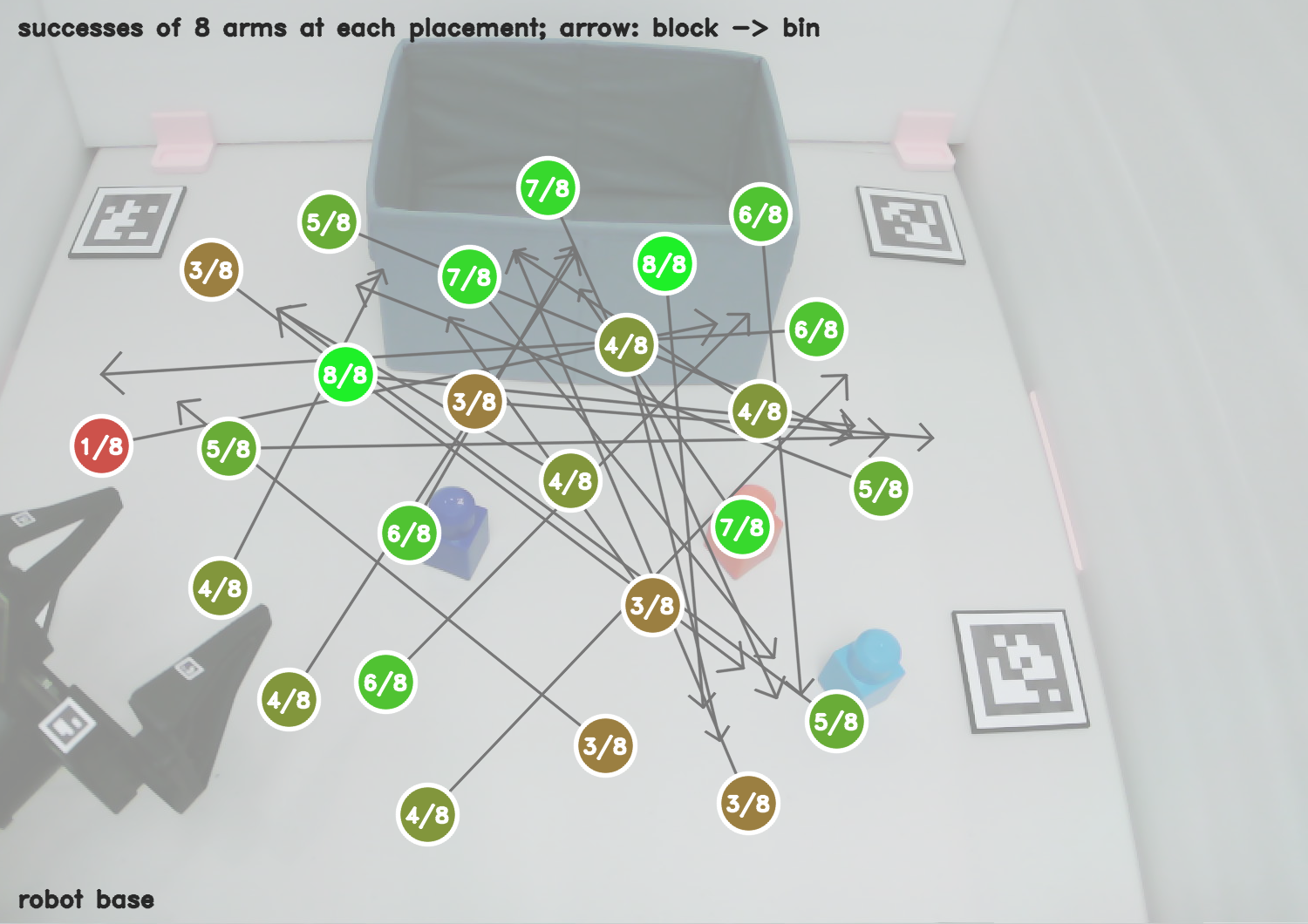}
  \caption{The 25 placements over the fixed camera's view. Each disc is the successes of the eight arms at that placement; the arrow points to the scene's bin.}
  \label{fig:placements}
\end{figure}

\begin{figure}[h]
  \centering
  \includegraphics[width=\linewidth]{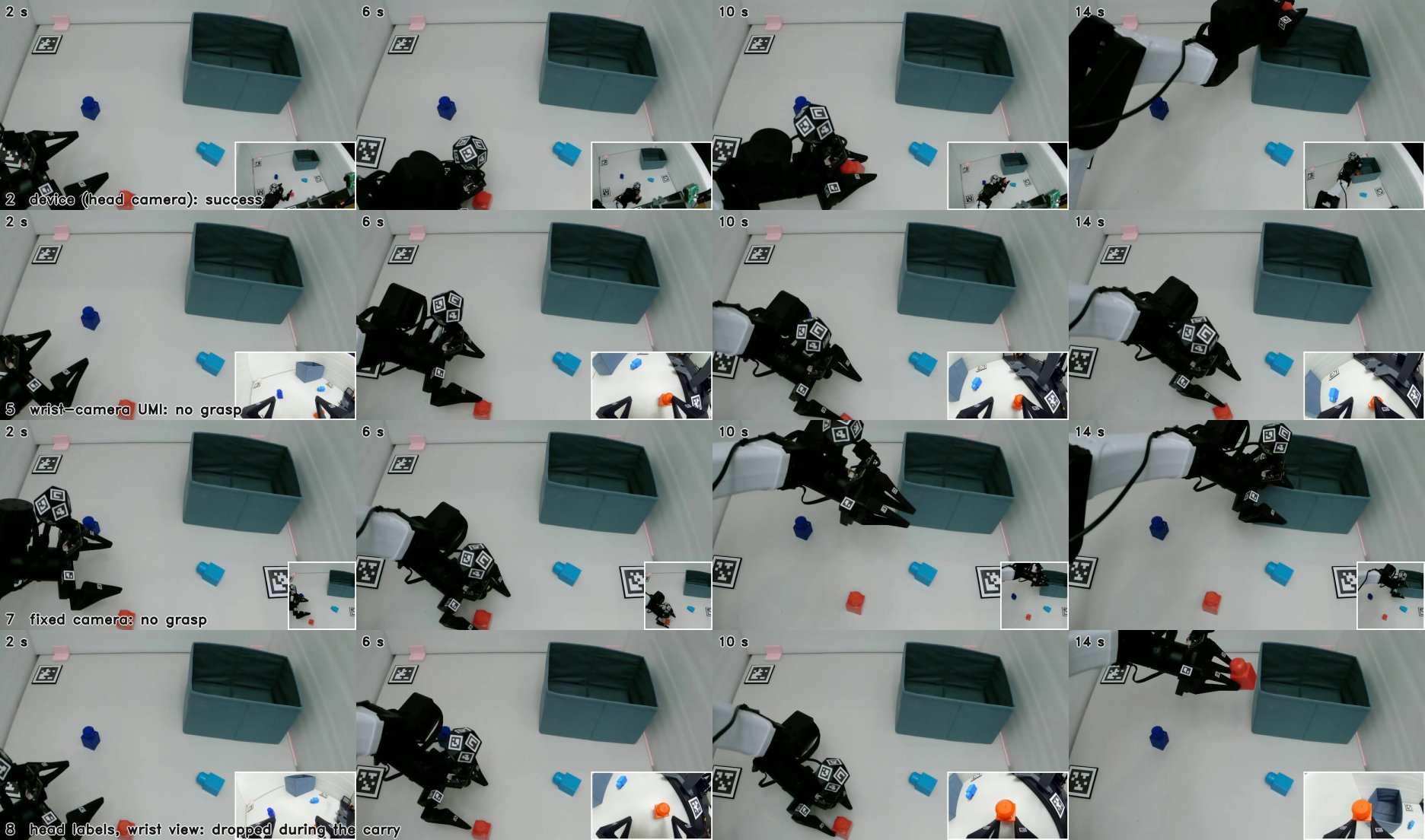}
  \caption{One placement (scene 16), the four camera arms, at 2, 6, 10 and \SI{14}{s}, from the fixed workspace camera with each policy's own view inset. During the carry the wrist view (rows 2 and 4) holds the block and the far wall and nothing else; the head view (row 1) keeps gripper and bin in one frame. Arm 8 opened with the bin out of its view.}
  \label{fig:rollouts}
\end{figure}

\end{document}